\documentclass[letterpaper, 10 pt, conference]{ieeeconf}  

\IEEEoverridecommandlockouts                              

\usepackage{graphicx} 
\usepackage{times} 
\usepackage{amsmath} 
\usepackage{amssymb}  
\usepackage{algorithm}
\usepackage{algpseudocode}
\usepackage{adjustbox}
\usepackage{booktabs}
\usepackage{soul}
\usepackage{xcolor}
\usepackage{tabularx}

\title{\LARGE \bf
Component Selection for Craft Assembly Tasks}

\author{Vitor Hideyo Isume$^{1}$, Takuya Kiyokawa$^{1}$, Natsuki Yamanobe$^{2}$, Yukiyasu Domae$^{2}$, Weiwei Wan$^{1}$\\ and Kensuke Harada$^{1, 2}$%
\thanks{© 2024 IEEE.  Personal use of this material is permitted.  Permission from IEEE must be obtained for all other uses, in any current or future media, including reprinting/republishing this material for advertising or promotional purposes, creating new collective works, for resale or redistribution to servers or lists, or reuse of any copyrighted component of this work in other works.}
\thanks{*This paper is based on results obtained from a project, JPNP20006, commissioned by the New Energy and Industrial Technology Development Organization (NEDO).}%
\thanks{$^{1}$Vitor Hideyo Isume, Takuya Kiyokawa, Weiwei Wan and Kensuke Harada are with the Graduate School of Engineering Science,
Osaka University, Osaka, Japan {\tt\small isume@hlab.sys.es.osaka-u.ac.jp, \{kiyokawa, wan, harada\}@sys.es.osaka-u.ac.jp}}%
\thanks{$^{2}$Natsuki Yamanobe, Yukiyasu Domae and Kensuke Harada are with the National Institute of Advanced Industrial Science and Technology (AIST), Tokyo, Japan {\tt\small \{n-yamanobe, domae.yukiyasu\}@aist.go.jp}}%
}

\begin{document}

\maketitle
\thispagestyle{empty}
\pagestyle{empty}

\begin{abstract}

Inspired by traditional handmade crafts, where a person improvises assemblies based on the available objects, we formally introduce the Craft Assembly Task. It is a robotic assembly task that involves building an accurate representation of a given target object using the available objects, which do not directly correspond to its parts. In this work, we focus on selecting the subset of available objects for the final craft, when the given input is an RGB image of the target in the wild. We use a mask segmentation neural network to identify visible parts, followed by retrieving labeled template meshes. These meshes undergo pose optimization to determine the most suitable template. Then, we propose to simplify the parts of the transformed template mesh to primitive shapes like cuboids or cylinders. Finally, we design a search algorithm to find correspondences in the scene based on local and global proportions. We develop baselines for comparison that consider all possible combinations, and choose the highest scoring combination for common metrics used in foreground maps and mask accuracy. Our approach achieves comparable results to the baselines for two different scenes, and we show qualitative results for an implementation in a real-world scenario.
\end{abstract}

\section{INTRODUCTION}

\begin{figure}
    \centering
    \includegraphics[width=\columnwidth]{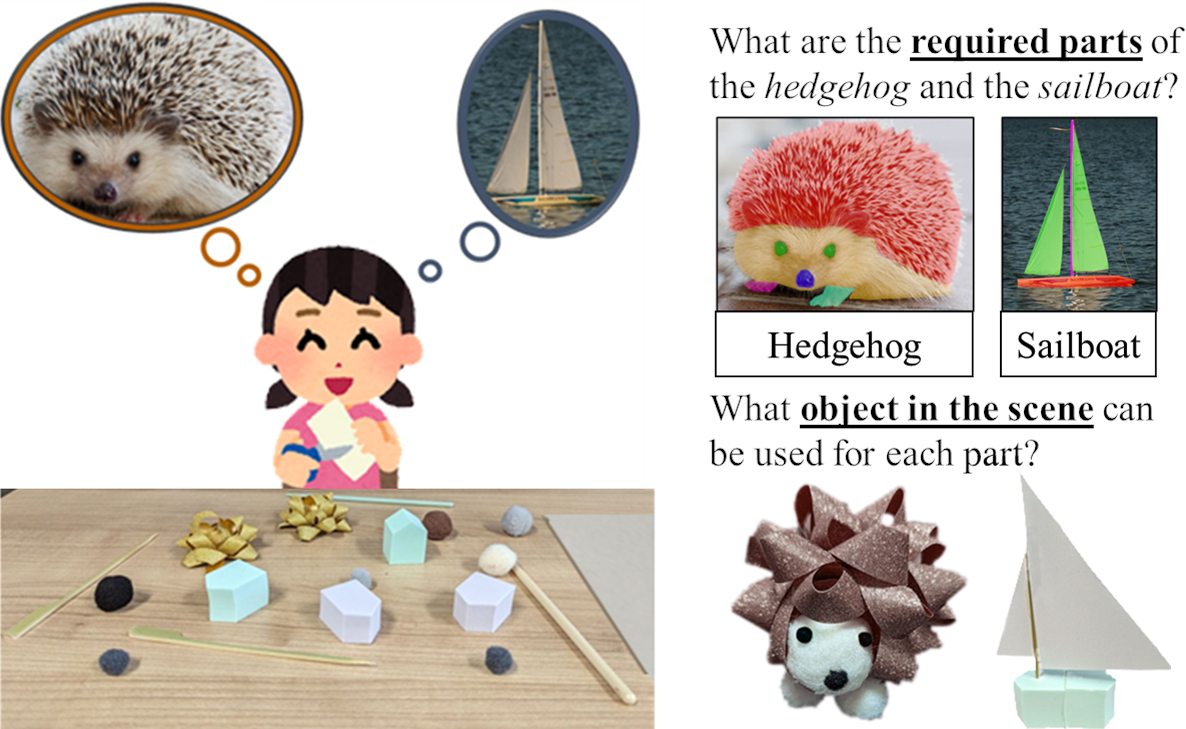}
    \caption{Illustrative example of a traditional craft task. Even if the available objects are non-exact correspondences to the target, a human is capable of abstracting and manipulating these objects to obtain a craft similar to their target.}
    \label{fig:example_intro}
\end{figure}

DIY (Do It Yourself) tasks aim to create, modify or repair objects using one's own ability and creativity without aid of professionals~\cite{kuznetsov2010rise}, covering a broad range of interests, such as woodwork, knitting, home-improvement and handmade crafts of objects. For instance, in arts classes, children are given simple materials such as colored papers, cotton balls, PET bottles, and are tasked with building a specific object without explicit instructions. This process requires a series of decisions, such as determining the necessary parts of the target object, and which materials are more suitable for each part. For example, consider the scenario illustrated in Fig.~\ref{fig:example_intro}, where the target objects are a hedgehog and a sailboat.

Despite their differences, both can be crafted with the available objects, demonstrating the versatility and creativity involved in these tasks. In addition, desired affordances, such as \textit{"floating"} for the sailboat craft can be taken into account; in this case, makeup sponges can be used as the hull since they are capable of floating. While these decisions may be straightforward for a human, automating them is challenging as they often rely on abstracting the available materials, evaluating possible interactions and prior knowledge of the target object. 

These challenges inspired us to propose the Craft Assembly Task: a novel robotic assembly task where the objective is to construct a craft that provides an accurate and functional representation of a given target object using a set of available objects that do not directly correspond to its parts. The inherent open-ended nature of the Craft Assembly Task allows us to explore and address object assembly's challenges when the available materials are limited or the specifications for the parts of the target object are ill-defined.

In our previous work~\cite{isume2021using}, we briefly introduced an early version of this task, where the input was a CAD model of the target assembly, and only focused on comparing the parts with objects in the scene. In this work, we expand upon and formally define this task, considering a more complex scenario where the input of the system is a single RGB image of the target object in the wild and a set of available objects. The expected output is a part-segmented 3D structure of the target that uses the available objects as its components. 

Our approach to this task begins by identifying the visible parts of the target object in the RGB image using a neural network. To reconstruct their 3D structure, we utilize template meshes, which are pre-prepared meshes used to represent object classes, due to a lack of databases with ground truth correspondences between real world RGB images and part-segmented 3D models. The template meshes corresponding to the detected object class are retrieved, and then undergo pose optimization, to identify the template mesh and its respective transformation that best aligns it to the input. To further minimize differences, only the parts of mesh that have corresponding masks are retained, and the occluded parts are generated using heuristics, such as symmetry assumptions. To compare the generated model parts with the scene objects, both are simplified to simple primitive shapes, such as cuboids or cylinders. The 3D bounding box of each part is used to generate the primitive candidates, and the best matching one is selected. Finally, we propose a search algorithm to match each simplified part with a scene object based on the local proportion between the pair, and the overall proportion of the whole model. The final output is a set of scene objects suitable for assembling a craft that matches the target object in appearance and functionality.

In this work, we do not consider the assembly sequence planning step. For evaluation, we apply the 6D transformations from the simplified parts to the chosen scene objects to obtain an assembled version in a virtual environment. As there is no ground truth solution for this task, we propose to evaluate the success rate of our task based on different metrics to account for accurate 3D pose, correct amount of parts and silhouette matching with the input. We also compare the part segmented rendering of our final craft with baselines, which analyzes all possible combinations of the scene objects and chooses the optimal one according to different metrics used in foreground maps and mask accuracy evaluation. Our approach is applied to four object classes under two different scenes, and it achieves comparable results to the baselines in the average part IoU (Intersect over Union) metric.

In summary, our main contributions are: 
\begin{itemize}
    \item Formally introducing the Craft Assembly Task, a challenging open-ended assembly scenario, inspired by DIY handmade crafts;
    \item Developing a framework to address this task by using template mesh retrieval, which alleviates the requirement of extensive part-segmented 3D models of the target objects;
    \item Proposing a search algorithm to find the most similar correspondence for cuboids and cylinders shapes, particularly when exact matches are unavailable, based on proportions of the dimensions.
\end{itemize}

\section{Related Work}

\subsection{3D reconstruction from RGB images}
Reconstructing a 3D model from a single or multi-view RGB input is a major topic of interest in computer vision. Most works rely on supervised deep learning approaches to find a direct correlation between 2D and 3D features, as well as dealing with occluded parts. However, initially there was a lack of large datasets for this task. To circumvent this limitation, Su et al.~\cite{Su2015} proposed to leverage the existing large scale 3D models dataset, ShapeNet\cite{chang2015shapenet}, by rendering its 3D models under different viewpoints with randomized backgrounds, obtaining correspondences between the transformed 3D model and the RGB image for training and testing. This was used in subsequent approaches~\cite{Choy2016,Wang2018}, although there was concern about performance when applied to real-world images, instead of rendered models. Xie et al.~\cite{Xie2020} used a similar approach, training on 3D models from ShapeNet rendered in 3D environments, and reported results on the Pix3D dataset~\cite{Sun2018}, a more challenging dataset which provides fine-aligned pairs between real-world images and 3D models of objects. Although it achieved state-of-the-art performance, there is a significant difference compared to rendered objects.

A common limitation in these methods is the required dataset for training, as most of them use hundreds of models per object class to account for small variations within the same object class. In the Craft Assembly Task, since the final craft features are limited by the available materials, our proposed solution focus on a 3D template mesh retrieval approach, as the main priority is recovering the overall segmented geometric structure and not a fine-grained aligned 3D model. This also alleviates the data required to retrieve a feasible structure.

\subsection{Recovering 3D structure from an RGB image}
For a craft assembly, reconstructing the whole unsegmented model in 3D does not provide sufficient information. We also need to retrieve the structure of the object: the set of parts and necessary transformations. Since datasets with segmented 3D models were scarce, Xu et al.~\cite{Xu2018} used a series of heuristics to segment the models from ShapeNet into a set of cuboid shapes, which were then used for training their neural network. With the release of PartNet~\cite{mo2019partnet}, a large-scale part-segmented 3D object dataset with different levels of granularity, works such as PQ-Net~\cite{wu2020pq} and SM-Net~\cite{yu2021sm}, used autoencoders to learn reconstruction of fine-grained parts, however the number of parts to be recovered must be specified beforehand.

Alternatively, unsupervised approaches proposed to learn free-form sets of primitives as parts to reconstruct the object~\cite{Paschalidou2020,Yao2021}. They also require specifying the number of possible parts beforehand to handle occlusion and avoid over-fitting with many small primitives. For the Craft Assembly Task, the primitives generated by the aforementioned unsupervised approaches may not be suitable for assembly, while in supervised approaches, there is a burden of generating a new dataset with the desired level of segmentation. Because our solution uses 3D template mesh retrieval, in order to obtain a segmented model, we created colored textures to label the parts of the meshes. It allows, during pose optimization between the template meshes and the input, for the final rendered image of the template mesh to already be part segmented. This segmentation helps distinguishing poses with ambiguous silhouettes.

\subsection{Part Assembly}
The next consideration to take into account is how to compare the parts of the transformed template mesh with the available parts in the scene. Several researches~\cite{Li2020,zhan2020generative} focused on the pose estimation sub-problem of autonomous assembly, where given the target assembly and point clouds of the parts, it learns the 6D-pose of each part to form the target assembly. It slightly differs from the Craft Assembly Task since, for the latter, the parts are non-exact correspondences. The closest work, to our knowledge, is general part assembly~\cite{Li2023}, where they learn to segment the target point cloud jointly with pose estimation of the input point clouds of the parts. They also focus on novel assemblies, by augmenting their training data with non-exact parts, however, in the Craft Assembly Task, we do not explicitly specify which of the objects from the scene should be used to form the assembly beforehand, adding another layer of decisions to the task.

\section{Methodology}
The Craft Assembly Task is formally defined as follows: given a representation of the target object and a scene with $n$ objects available, which are non-exact correspondences to the required parts, the goal is to select a subset of $m$ objects from the scene and assemble a craft that accurately represents the target object in terms of both appearance and functionality. We focus on the object selection process, particularly the scenario where the representation of the target object is a single RGB image of the object in the wild. Furthermore, we only consider scenes where the available objects are primitive shapes, either cuboids or cylinders.

\subsection{Overview}
\begin{figure*}[t]
    \centering
    \includegraphics[width=\linewidth]{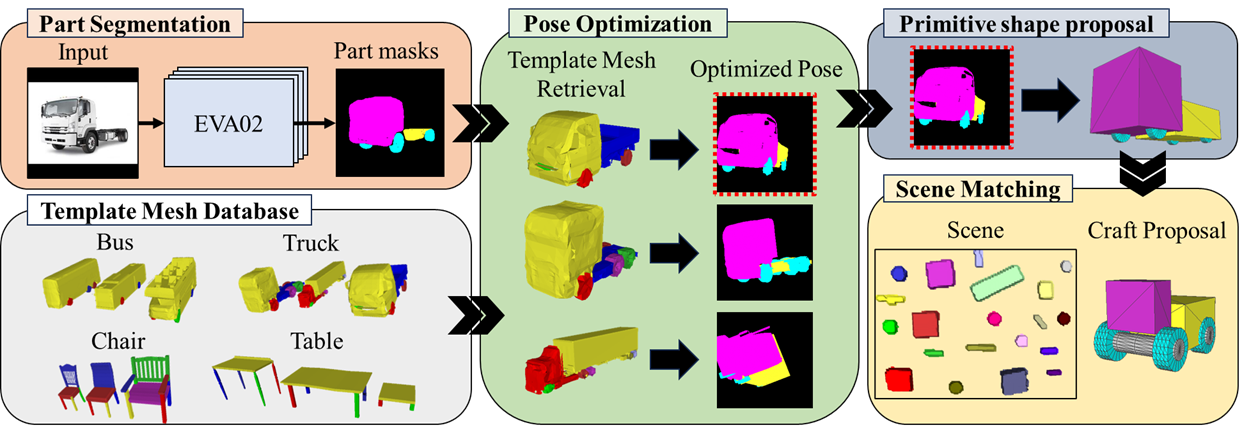}
    \caption{Overview of our proposed solution for the Craft Assembly Task. Given the RGB image of the target object, the visible parts are segmented and classified. Labeled template meshes of the detected class are retrieved from a prepared database and their pose are optimized through a differentiable renderer using the segmentation results as the target. The parts of the best aligned mesh are simplified to primitive shapes. Finally each part is matched with an object in the scene (input) using a search algorithm, generating the final Craft Proposal.}
    \label{fig:overview}
\end{figure*}

There are two main challenges: obtaining a part-segmented 3D structure suitable for assembly from a single RGB image, and matching the parts of this structure with the objects available in the scene. Our approach to address these challenges are divided in four steps, which are illustrated in Fig.~\ref{fig:overview}. In the first step, we obtain the part segmentation masks from the RGB input using a fine-tuned vision transformer, in this case, EVA02~\cite{fang2023eva}.

The second step is retrieving the template mesh from a database alongside the optimized camera parameters that best fit the masks obtained in the first step. These template meshes, representing the considered object classes in 3D, are prepared beforehand, using texture maps to label their parts. Using a differentiable renderer, we perform online training to optimize the camera parameters for each template mesh of the detected object class, represented as the "Pose Optimization" block in Fig.~\ref{fig:overview} . We follow a training approach similar to~\cite{Pavllo2021}, initializing $N_c$ initial views per batch, and optimizing simultaneously using silhouette loss. It results in $N_c$ images per batch, which are individually scored  using a weighted sum of three loss terms: the IoU loss, the part IoU loss and a loss based on the distance between centers of parts. The result with the minimum total loss is selected for the next step.

In the third step, we refine the result by considering only the parts from the retrieved template mesh that have corresponding masks from the first step. To generate the remaining parts of the model that were occluded in the input, we assume left-right symmetry, and apply rules designed for specific object classes to add internal components that are unseen in both the input and the template meshes. It ensures functional coherence, for example, adding an axle to connect wheels so they afford to "roll together".

Comparing the parts of this generated model with the scene objects is non-trivial due to non-exact correspondences. Traditional 3D measures, such as chamfer distance, are inadequate due to scale differences and may not guarantee visual likeness. Therefore, we propose to first simplify each part of the model to the same domain of the scene objects: primitive shapes, either a cuboid or cylinder, by using its 3D bounding box to propose primitive candidates. We evaluate the primitive shape candidates by sampling a point cloud for each and calculating the chamfer distance to a sampled point cloud of the part, choosing the candidate with lowest distance.

Finally, in the last step, a search algorithm matches each part in the simplified primitive-shaped model with the closest correspondence in the scene, considering both per-part proportion, that is, the proportions between the dimensions of the part, and the overall proportion, the proportion of the dimensions between different parts. Our strategy revolves around using the largest part in the model as the reference of all other parts, and then finding its correspondence in the scene. The corresponding scene object is used as the reference to all other scene objects, obtaining a common ground for comparison.

\subsection{Part segmentation} 
To perform part segmentation suitable for assembly, we label images extracted from ImageNet\cite{Deng2009} to fine-tune a pre-trained EVA02 model. Specifically, we use the weights pretrained on the COCO dataset\cite{lin2014microsoft} with Object365\cite{shao2019objects365} intermediate fine-tuning, and fine-tune it for the parts of the truck, chair, table and bus object classes, resulting in nine possible part classes. They are listed alongside their corresponding object classes in Table \ref{mask_labels}. To minimize errors in part mask classification, due to similar object parts and objects in the background, we use the EVA02 model with pre-trained weights for object detection, specifically the COCO with Object 365 intermediate fine-tuning weights, as an off-the-shell model to crop the target object and classify it in the considered object classes. In cases multiple objects are detected, we choose the one with the highest confidence. In case the object detection fails, the mask segmentation is applied to the whole image. Since some part classes are unique to a object class, for example, \textit{"truck\_cabin"} is exclusive to object class \textit{"truck"}, this correlation can be used to determine the object class. Finally, in the case multiple unique part classes are detected, the one with the highest confidence is chosen to determine the object class.

Formally, for each input image $I$, with a single instance of a target object, we obtain a set of masks $M=\{m_1, m_2, ... , m_T\}$ and an object class label, based on the object detection results or the unique part class label present in $M$.

\begin{table}
\small
\caption{Considered part classes with corresponding object classes.}
\label{mask_labels}
\begin{center}
\begin{tabular}{l|c c c c}
& \multicolumn{4}{c}{\textbf{Object classes}} \\
\hline
\textbf{Part classes} & truck & bus & chair & table\\
\hline
truck\_cabin & $\bigcirc$ & $\times$ & $\times$ & $\times$\\
truck\_body & $\bigcirc$ & $\times$ & $\times$ & $\times$\\
bus\_body & $\times$ & $\bigcirc$ & $\times$ & $\times$\\
wheel & $\bigcirc$ & $\bigcirc$ & $\times$ & $\times$\\
chair\_back & $\times$ & $\times$ & $\bigcirc$ & $\times$\\
chair\_seat & $\times$ & $\times$ & $\bigcirc$ & $\times$\\
chair\_arm & $\times$ & $\times$ & $\bigcirc$ & $\times$\\
table\_surface & $\times$ & $\times$ & $\times$ & $\bigcirc$\\
furniture\_leg & $\times$ & $\times$ & $\bigcirc$ & $\bigcirc$\\
\hline
\end{tabular}
\end{center}
\end{table}

\subsection{Template mesh Database}
To create the template mesh database, we collect three meshes per object class from freely available models at SketchFab~\cite{sketchfab} to act as templates. They are illustrated in Fig.~\ref{fig:overview}, and are stored in the database with semantic labels regarding their object class.

For each mesh, we create a custom texture map, labeling the faces according to the considered part classes. The meshes are scaled to fit within a unit cube, and are centered at the origin of the coordinate system.

\subsection{Pose optimization}
Using the results from the part segmentation step, we retrieve all template meshes of the corresponding object class. Our goal in this step is determining the camera parameters and which retrieved template mesh best aligns the rendered image with the masks in $M$.

For each retrieved template mesh, we perform pose optimization following the approach from~\cite{Pavllo2021} with some modifications. They initialize $N_c$ camera hypotheses per template mesh model per batch to avoid local optima issues. The virtual camera parameters of a differentiable renderer are optimized to match the silhouette of the render to a given target silhouette. The highest scoring results, in terms of IoU, are selected, and in the following step, they apply semantics to differentiate ambiguous poses, choosing the result with the highest average part IoU, that is the mean IoU calculated per-part based on the semantic labeled areas.

In our approach, we combine all masks from $M$ into a single binary mask, $M^*$,  which is used as the target silhouette during training. Additionally, we modify the loss function from mean squared error ($L_2$ loss) to mean absolute error ($L_1$ loss), as employing $L_1$ loss yielded more consistent results in pose estimation. Empirically, we observed that $L_2$ loss converges faster, but it resulted in incorrect poses for "table" and "chair" more often.

For the final selection from the $N_c \times batches$ results, relying solely on IoU or part IoU does not lead to accurate poses due to significant divergences between our template meshes and the input. Instead, we propose selecting the result that minimizes the weighted sum of three losses. The masks in $M$ are combined based on their part class label, generating the set $\{m_1^*, m_2^*, ..., m_O^*\}$, with each mask corresponding to a unique part class label. Each result is a render $R$ also divided into a set of masks with unique part class labels: $\{r_1, r_2, ... r_O\}$. In case a label is present in $M$ but not in $R$, the mask is considered empty for evaluation.

The first loss, $\mathcal{L}_{IoU}$, represents the IoU loss of the overall silhouette, aiming to ensure alignment of the overall structure, but being subject to ambiguous poses.
\begin{equation}
\mathcal{L}_{IoU} = 1 - IoU(M^*, R)
\end{equation}
The second loss, $\mathcal{L}_{mIoU}$, is the part IoU loss averaged across the $N$ semantic classes, which helps handling pose ambiguity by considering the accuracy of the semantic labels. However, due to differences between the template and the input, this loss can be high even if they are well-aligned.
\begin{equation}
\mathcal{L}_{mIoU} = \frac{1}{N} \times \sum_{i=1}^{N} 1-IoU(m_i^*, r_i)
\end{equation}
The third loss is a normalized Euclidean distance between the centers of masks averaged across the $N$ semantic classes. It alleviates the the penalties of the other losses, by considering only the alignment of the center of the masks:
\begin{equation}
\mathcal{L}_{dist} = \frac{1}{N} \times \sum_{i=1}^{N} d_{center}(m_i^*, r_i)
\end{equation}
In this equation, $d_{center}$ stands for the normalized Euclidean distance function between the center of the pair of masks. They are normalized by using the diagonal distance of the top-left corner to the bottom-right corner of the masks, which serves as the maximum possible distance. If one of the masks is empty, the maximum possible distance is adopted.

Each component's influence is controlled by a weight $\lambda$. Finally, we retrieve the result that minimizes the total loss $\mathcal{L}_{total}$.:
\begin{equation}
\mathcal{L}_{total} = \lambda_{IoU} \times \mathcal{L}_{IoU} + \lambda_{mIoU} \times \mathcal{L}_{mIoU} + \lambda_{dist} \times \mathcal{L}_{dist}
\end{equation}

\subsection{Primitive proposal}
In this step, we begin by addressing differences in the number of parts between the retrieved template mesh and the input. For example, if the input has a truck with four wheels, but in the chosen template mesh, the truck has six wheels. After rendering the result from the pose optimization step, we filter out parts of the template mesh that lack a corresponding mask in $M$. This correspondence is found by calculating the euclidean distance between each visible part in the rendered template mesh and the masks in $M$.

To generate the missing components, we make two key assumptions. First, we assume left-right symmetry of the target object to generate the occluded parts. If a part has all vertices on either the left or the right side relative to the center line of the original template mesh, and there is no corresponding retrieved part on the opposite side, that part is cloned and mirrored. The second assumption is regarding internal components for certain objects where specific affordances are desirable. These internal components are usually absent in the images of the target object and in the template meshes, requiring predefined rules to generate a simplified version of them. Specifically, for the \textit{truck} and \textit{bus} object categories, the part \textit{wheel} should be connected to another parallel \textit{wheel} via a new part: an \textit{axle}, since the affordance \textit{roll} is desirable to keep the functionality in the craft assembly. The proposed \textit{axle} is a cylinder with sufficient length to connect the center of a pair of \textit{wheels}, and the radius is defined as 20\% of the radius of the wheels. They are created after proposing the primitive shapes for the wheels.

Once all parts of this adjusted model are obtained, each one is simplified to a single type of primitive shape, either a cuboid or a cylinder. Each part is aligned to the axis of the coordinate system to approximate the minimal bounding box as the axis-aligned bounding box. Four primitive shapes candidates are generated based on the dimensions of this bounding box: a cuboid shape, or three possible rotated cylinders, where the planar faces are aligned with one of the three possible directions. Then, we sample separate uniform point clouds for each primitive shape candidate and for the part to calculate the chamfer distance. The candidate with the lowest distance is chosen as the simplified representation of that part. Following this process to all parts, a primitive shape model is obtained.

\subsection{Scene matching}
To compare the parts from the primitive shaped model with the objects available in the scene, which are also considered to be primitives, we designed a search algorithm, it aims to find, for each part, an object in the scene that has the same primitive shape type and that best matches the proportions of that part, while also considering the overall proportions between different parts. Since the parts in the model and the objects in the scene don't have the same scale, we first try to find a single correspondence and then calculate the relative dimensions of all other parts compared with this first correspondence to ensure the proportions are kept locally and globally. The algorithm is divided into three steps: 1) finding the part with the largest dimensions in the model; 2) finding the object in the scene that corresponds with the part from step 1; 3) finding all remaining correspondences.

In the first step, all the parts from the primitive shaped model $P$ are analyzed in order to find the part with the single largest dimension, $p_{max}$. Then all dimensions of all parts from $P$ are normalized using the largest dimension from $p_{max}$, generating a set of normalized parts, $P^N$. Similarly, we need to normalize the dimensions of the objects in the scene according to a single object, to make a direct comparison with the parts in $P^N$. 

In the second step, given the scene $S$, each object $o_k$ in the scene $S$ is normalized using its own largest dimension, generating the set $S^*$. Then, we aim to find the normalized object that best matches the normalized $p_{max}$ ($p^N_{max}$) by ranking them using the average absolute difference between dimensions, where a lower difference corresponds to a higher ranking. To avoid cases where the highest ranking object is small compared to other objects in the scene, thus hindering the selection of the following objects as they should be smaller than this first object to keep the global proportion of the craft, we also rank the objects in the scene according to their volume, where a higher volume results in a higher ranking. The object correspondence, $o_{max}$, for $p^N_{max}$ is chosen based on the highest average placement in both of these rankings.

Finally, in the third step, all objects in $S$ are normalized using the largest dimension of $o_{max}$, obtaining the set $S^N$. For each remaining part $p_i^N$ in $P^N$, we evaluate each normalized scene object $o_j^N$ in $S^N$ based on the difference between their dimensions, $\epsilon_{dim}$, and the difference between their ratios, $\epsilon_{ratio}$. The former, $\epsilon_{dim}$, is calculated by averaging the absolute differences between the dimensions of $p_i^N$ and $o_j^N$. The error $\epsilon_{ratio}$ is calculated by averaging the absolute difference between the ratios of $p_i^N$ and $o_j^N$. The ratios are computed by dividing each dimension by the maximum dimension of their respective part or object. This comparison ensures the overall proportions are maintained, which is particularly important for cylinder shapes to retain the proportion between the radius and length.

The final total error is then calculated as the square root of the sum of the squared errors. The scene object that minimizes the total error and has not been selected yet is chosen to represent the part $p_i^N$ in the final craft. This process concludes when a unique corresponding object in the scene is found for each part.

\section{Evaluation}

\subsection{Implementation details}
To fine-tune the pre-trained EVA02 model, we collected images from ImageNet and remove instances where the target object is excessively small, occluded or computer-rendered. The final dataset has 1193 images, manually annotated for the part classes listed in Table~\ref{mask_labels}, with 945 images allocated for training. It is trained for 40k steps, with the image size set to 512x512 pixels and batch size to 2. All other parameters are left at their default values. For evaluation, both object detection and mask segmentation confidence thresholds are set to 0.75.

In the Pose Optimization step, each retrieved template mesh is trained for 100 steps, with 40 initial views, with a learning rate of 0.1. To further avoid local optima issues, five batches for each template mesh are trained simultaneously. For the total loss evaluation, from Eq.~4, we set $\lambda_{IoU} = 0.75$, $\lambda_{mIoU} = 0.15$ and $\lambda_{dist} = 0.15$, manually optimized according to the results for a small validation set.

Regarding the available objects in the scene, we assume their shape and dimensions are known. The considered objects have either a cuboid or cylinder shape. Then, two scenes are generated:

\textbf{Scene I}: a diverse scene with 20 types of primitives, each with 10 instances, for a total of 200 available objects.

\textbf{Scene II}: a more restricted scene with 10 types of primitive, with a total of 20 objects. The dimensions and shapes are collected from real-life objects such as cardboard boxes, plastic tubes, wood cylinders etc.

\subsection{Metrics}
In the Craft Assembly Task, the final craft depends on the available objects in the scene, leading to different results for different scenes even if the input image is the same. Therefore no ground truth solution is available for direct comparison. To analyze the success rate~(SR) of our method, we consider three different metrics. First, for 3D alignment, we evaluate the viewpoint accuracy~\cite{tulsiani2015viewpoints} (VP Acc) at 30\textdegree. Then, we consider if the number of instances for each part class label correspond to the ground truth~(Part Acc). Finally, we evaluate the 2D alignment by checking if the silhouette IoU is above a threshold~(Sil. Acc), here set to 0.5, since they will inherently be different due to the non-exact correspondence nature of the task. If our proposed craft satisfies these three metrics, we count it as a successful proposal, otherwise we consider it as a failure.

Additionally, to further analyze the similarity with the input image, we evaluate the part IoU. In this case, we render our craft with the ground truth pose, to ensure the comparison is fair. We compare these results against a heuristic approach, which is used to generate two baselines. This heuristic is provided with the ground truth part segmentation annotations, the objects in the scene, annotated camera poses, and the exact number of parts for the target object in the input images. It explores all possible combinations of objects, pruning those where the shape of the selected object and the part don't match, or violate specific object classes rules, such as: "all \textit{furniture legs} of a \textit{chair} or a \textit{table} must have the same dimensions" and "all \textit{wheels} of a \textit{truck} or a \textit{bus} must have the same dimensions". For each of the remaining combinations, the objects are positioned using pre-defined rules and are rendered using the annotated camera poses, resulting in mask-segmented renderings such as the outputs of the pose optimization step. Then, they are evaluated using two metrics, generating two baselines:

\textbf{mIoU baseline}: Average mask IoU across all parts classes present in the ground truth annotation.

\boldmath
$mE_{max}$ \textbf{baseline}: E-measure~\cite{Fan2018}, metric used in foreground masks evaluation, calculated per-part class masks and averaged.
\unboldmath

These baselines select combinations that are directly optimized for 2D alignment with the input, assuming the pose and amount of parts are correct. To mitigate localization issues, the area inside the bounding box of the rendered object and the ground truth annotations are cropped and extended to a 256x256 pixels size during evaluation, preserving the aspect ratio with padded black areas.

\subsection{Experimental results}
\begin{figure}[!t]
    \centering
    \includegraphics[width=\linewidth]{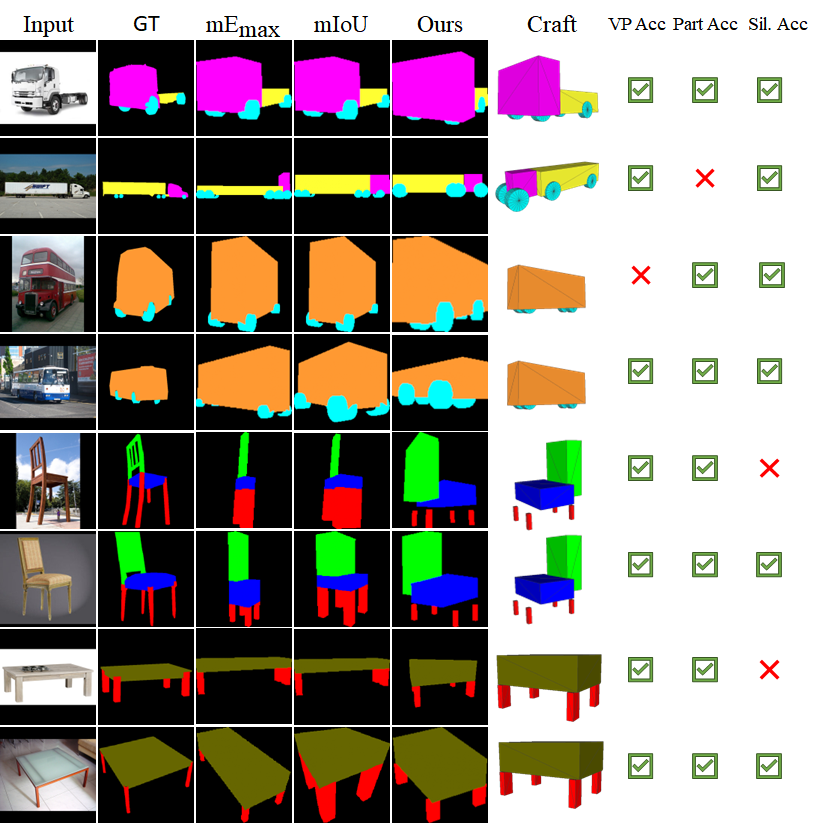}
    \caption{Renders of the final combinations for the baselines and our method. We also show the evaluation results for each metric for our proposed craft.}
    \label{fig:qualitative}
\end{figure}

Qualitative results of both our method and the baselines for Scene I are illustrated in Fig.~\ref{fig:qualitative}, alongside the evaluation of the metrics for determining if the craft is successful. Among the baselines, mIoU appear to better maintain the overall part proportions, while $mE_{max}$ tend to over-fit a few parts. Our method also seems to preserve overall proportions, however, although we allow template mesh deformation during the pose optimization step, it doesn't produce significant variation in the primitive-shaped model, resulting in most parts proportions being the same as the original template mesh.

For quantitative evaluation, we report the results for the success rate and the part IoU comparison in Table~\ref{tab:success}. For Scene I, we evaluated 248 images withheld from training, while in Scene II, 193 images are evaluated. The variation of images used for each scene evaluation is caused by limitations in the available objects in Scene II. Specifically, some instances in the \textit{truck} object class need 6 or more wheels, however there aren't sufficient objects in the scene that could act as wheels. Consequently, such cases are excluded.

Regarding the success rate evaluation, in the VP Acc evaluation of the \textit{bus} and \textit{table}, we take into account their symmetry, as most of their models don't have distinguishing features between the front or back of the object. Most object classes presented failure in one of the metrics. For \textit{truck}, although the pose and overall silhouette are often correct, a typical point of failure is retrieving all wheels from the input during the segmentation step. This is mainly observed in images where the \textit{truck} has 8 or more \textit{wheels}, as we can observe in the significant increase in Part Acc in Scene II, after removing such images from the evaluation due to lack of available objects. For \textit{chair}, it typically fails in the Sil. Acc, where our craft often has shorter \textit{furniture\_leg} compared to the input, causing poor alignment of the silhouettes. For \textit{table}, it also fails in the Sil. Acc, in this case, the \textit{table\_surface} being shorter in length than the input causes the misalignment. Finally, for \textit{bus}, we achieve a high success rate, possibly due to the simplicity in shape and overall proportions not varying as much.

For the part IoU, the mIoU baseline should achieve the highest possible score since it is directly optimized for this metric, acting as an approximation of an optimal solution. To be able to directly compare 2D similarity, we render our results with the ground truth pose, reporting it as \textit{"Ours w/ gt"}. In both scenes, our method achieves comparable scores to the $mE_{max}$ baseline, but under performed when compared to the mIoU baseline. We believe the issues observed in the success rate analysis, specifically about the Sil. Acc, are amplified here since the alignment is conducted per-part class. A possible future improvement is allowing per-part deformation during or after pose optimization, to fine-tune the dimensions of the parts, making the system more adaptable without increasing the number of template meshes.

\begin{figure}[!b]
    \centering
    \includegraphics[width=0.9\columnwidth]{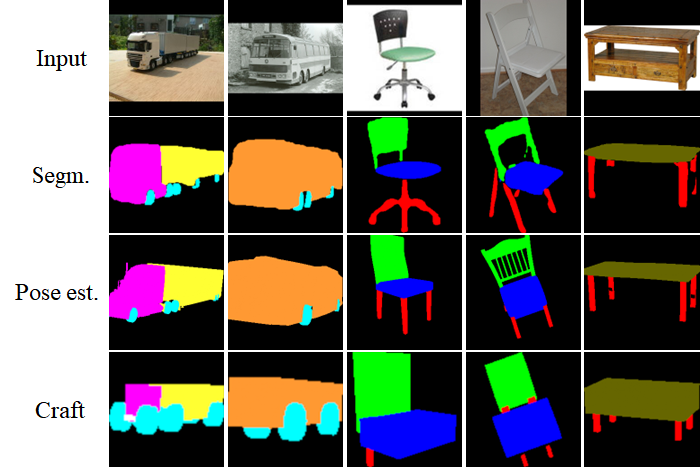}
    \caption{Qualitative examples of our method applied to novel instances. Although the final craft cannot match the input, due to structural differences in our template meshes, the segmentation and pose can still be retrieved.}
    \label{fig:novel}
\end{figure}

We also show some qualitative examples of our method applied to novel instances that do not match the template meshes in Fig.~\ref{fig:novel}. In the case of \textit{truck} and \textit{bus}, the main difference is in the \textit{wheels} amount and positioning, respectively. As for the \textit{chair} examples, their legs structure is differs significantly from the template meshes. For the \textit{table} example, it includes a part (drawer at the bottom) that is not considered in our original dataset. Overall, the part segmentation is accurate, but the pose estimation fails in cases where the structure of the object is significantly different from the meshes, such as the case where the  \textit{chair} legs are at an angle, generating a very distinct silhouette.

\begin{table}[t]
    \centering
    \caption{Success rate considering multiple metrics and average part IoU evaluation of our method and baselines.}
    \resizebox{\linewidth}{!}{%
    \begin{tabular}{l|c c c c |c c c}
    \hline
    & \multicolumn{7}{c}{\textbf{SCENE I}} \\
    \hline
    & \multicolumn{4}{c}{Success Rate} & \multicolumn{3}{c}{Part IoU} \\
    \hline
    Class & VP Acc & Part Acc & Sil. Acc & SR &  $mE_{max}$ & mIoU & Ours w/ gt  \\
    \hline
    Truck & 0.726 & 0.258 & 0.935 & 0.177 & 0.273 & 0.427 & 0.262 \\
    Bus & 0.939 & 0.949 & 1 & 0.899 & 0.399 & 0.434 & 0.368\\
    Chair & 0.676 & 0.794 & 0.442 & 0.265 & 0.199 & 0.377 & 0.251 \\
    Table & 0.731 & 0.865 & 0.654 & 0.481 & 0.385 & 0.456 & 0.288\\
    \hline
    \textbf{Average} & 0.806 & 0.737 & 0.834 & 0.543 & 0.337 & 0.429 & 0.309 \\
    \hline
    \hline
    & \multicolumn{7}{c}{\textbf{SCENE II}} \\
    \hline
    & \multicolumn{4}{c}{Success Rate} & \multicolumn{3}{c}{Part IoU} \\
    \hline
   Class & VP Acc & Part Acc & Sil. Acc & SR &  $mE_{max}$ & mIoU & Ours w/ gt \\
    \hline
    Truck & 0.500 & 0.875 & 0.625 & 0.375 & 0.178 & 0.315 & 0.203 \\
    Bus & 0.939 & 0.949 & 1.0 & 0.899 & 0.279 & 0.281 & 0.254 \\
    Chair & 0.676 & 0.794 & 0.529 & 0.323 & 0.184 & 0.322 & 0.307 \\
    Table & 0.731 & 0.865 & 0.461 & 0.288 & 0.318 & 0.322 & 0.275 \\
    \hline
    \textbf{Average} & 0.819 & 0.896 & 0.756 & 0.611 & 0.269 & 0.300 & 0.267 \\
    \hline
    \end{tabular}}
    \label{tab:success}
\end{table}

\subsection{Real world implementation}
\begin{figure}[tb]
    \centering
    \includegraphics[width=\columnwidth]{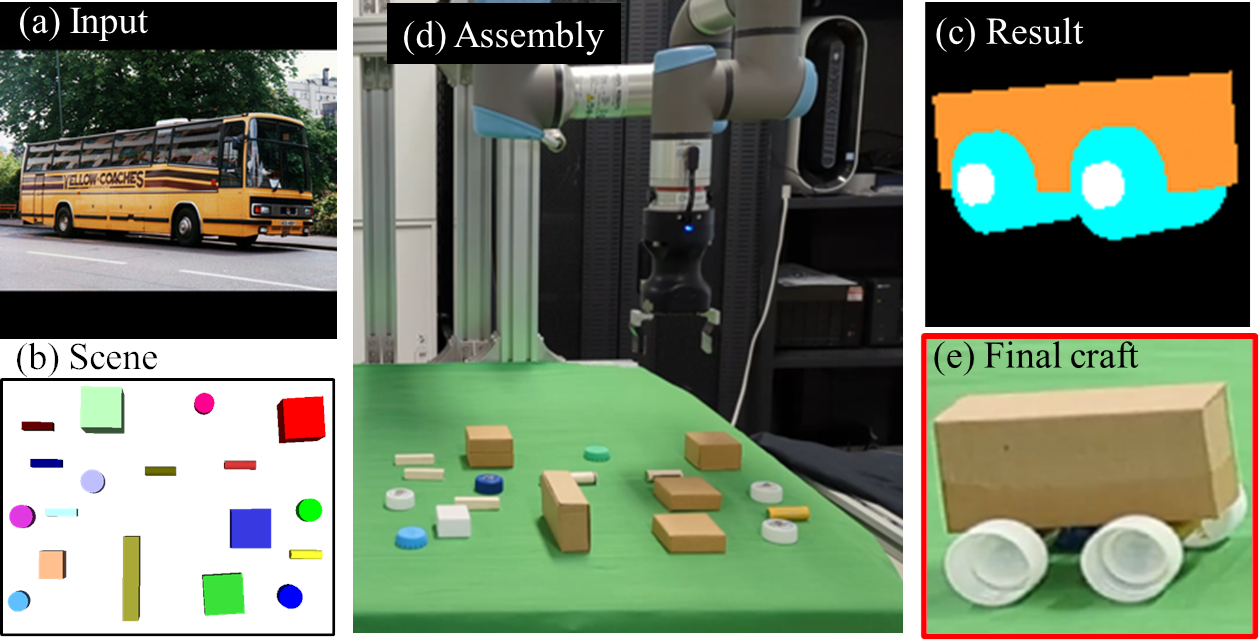}
    \caption{Example of implementation in a real world environment for the input image of a bus. During assembly, some adjustments to avoid collision are done through pre-defined rules. The render area is maximized for better visualization. A video demonstration for this input is available at: https://youtu.be/tjz2d\_NuxB8}
    \label{fig:test2}
\end{figure}

We show an implementation of the system in a real world setting, using two UR3e arms for manipulation, and some objects from Scene II as the available scene objects. The assembly motions are manually defined, and perception of the real world environment was captured beforehand using an Intel\textregistered \ RealSense\texttrademark \ SR305 camera. The qualitative result for the image of a bus in Scene II is shown in Fig.~\ref{fig:test2}.

\section{CONCLUSIONS}
Our proposed approach for the Craft Assembly Task offers robust solutions by utilizing template meshes to address the lack of ground truth data for this task. It achieved comparable performance to baselines that approximate optimal solutions. While our method effectively utilizes template meshes to reduce the need for extensive datasets, it has limited adaptability for object instances that substantially differ from the available templates. Scaling the proposed system to new object classes requires labeling part classes in images and template meshes. This could be facilitated by using automated segmented tools, such as Grounded-SAM~\cite{ren2024grounded}, for images, and PointNeXt~\cite{qian2022pointnext}, for point clouds of the meshes, as a starting point to, then, refine the annotations.

In our future work, we hope to address these limitations by exploring Large Language Models~(LLMs) for 3D reasoning. A possible extension towards a more general system would be using LLMs with 3D model generation capabilities~\cite{yin2023shapegpt} to first generate a 3D model from the RGB image and then using its reasoning capabilities to infer about the part segmentation and necessary internal components.

\section{ACKNOWLEDGMENTS}
Some of the 3D models used as template meshes in this work were created by maregajavier, PAndras, chirag.swamy.1993, oakar258, businessyuen, Daniel JIN, Intee Assets, Kemron, Nichgon and Poring and obtained under the Creative Commons Attribution (CC-by) license. The remaining models were created by vegu and VRC-IW and obtained under the Creative Commons Attribution-NonCommercial license. All models were collected from SketchFab.

\addtolength{\textheight}{-12cm}   





\bibliographystyle{IEEEtran}
\bibliography{IEEEabrv, all_ref.bib}
\end{document}